\documentclass[letterpaper,10pt,conference]{ieeeconf}
\usepackage{cite}
\usepackage{amsmath,amssymb}
\usepackage{graphicx}
\usepackage{wrapfig}
\usepackage[footnotesize]{subfigure}
\usepackage{float}
\usepackage{etoolbox}
\makeatletter
\pretocmd{\@subcaption}{\nointerlineskip}{}{%
  \PackageError{figure_style}{Could not tighten subcaption spacing}{}%
}
\makeatother
\newlength{\practicalfigurecontentwidth}

\usepackage{xcolor}
\usepackage[hidelinks,hyperfootnotes=false]{hyperref}
\hypersetup{pdfauthor={Younghyo Park, Antonia Bronars, Pulkit Agrawal},pdftitle={When Robots Outweigh Objects: How Inertia Mismatch Shapes Robot Policy Learning}}
\usepackage{mdframed}
\usepackage{needspace}
\usepackage{ragged2e}

\newcounter{mypar}[subsection]

\newcommand{\myparagraph}[1]{%
  \par\addvspace{0.05in}%
  \refstepcounter{mypar}%
  \noindent\textbf{#1}\space\ignorespaces
}

\newcounter{result}[subsection]
\renewcommand{\theresult}{\ifnum\value{subsection}=0\thesection\else\thesubsection\fi-\Roman{result}}
\newenvironment{myresult}[2][]{%
  \refstepcounter{result}%
  \ifx&#1&\else\label{#1}\fi%
  \begin{mdframed}[linewidth=0.5pt, linecolor=black, backgroundcolor=gray!5, nobreak=true]
  \textbf{Result~\theresult}~(#2):%
}{%
  \end{mdframed}%
}
\newcommand{\resultref}[1]{\hyperref[#1]{Result~\ref*{#1}}}

\newenvironment{insetparagraph}{%
  \par\addvspace{4pt}\noindent\begin{minipage}[t]{\columnwidth}%
  \setlength{\columnsep}{6pt}%
  \RaggedRight
}{%
  \par
  \ifnum\value{WF@wrappedlines}>2\relax
    \vspace{\dimexpr\value{WF@wrappedlines}\baselineskip-2\baselineskip\relax}%
  \fi
  \WFclear\end{minipage}\par\addvspace{6pt}%
}
\newenvironment{insetparagraph*}{%
  \par\addvspace{4pt}%
  \setlength{\intextsep}{3pt}%
  \setlength{\columnsep}{6pt}%
  \RaggedRight
}{%
  \par
  \ifnum\value{WF@wrappedlines}>1\relax
    \vspace{\dimexpr\value{WF@wrappedlines}\baselineskip-\baselineskip\relax}%
  \fi
  \WFclear\par\addvspace{6pt}%
}
\makeatletter
\newcommand{\insetfigurefits}{%
  \Needspace{\dimexpr\dp\WF@box+\baselineskip+\intextsep\relax}%
}
\makeatother
\graphicspath{{figures/}}
\begin{document}
\flushbottom
\title{\LARGE\bfseries When Robots Outweigh Objects:\\
How Inertia Mismatch Shapes Robot Policy Learning}
\author{%
\begin{tabular}{c@{\hspace{3em}}c@{\hspace{3em}}c}
Younghyo Park & Antonia Bronars & Pulkit Agrawal\\
 & MIT &
\end{tabular}}
\maketitle

\suppressfloats[t]
\begin{abstract}
Conventional robot arms can be extremely heavy compared with the everyday objects they manipulate, creating a significant inertia mismatch at contact. Even a small accidental bump from a heavy arm can knock a light object out of place. Modern learning methods have nevertheless demonstrated a wide range of successful manipulation behaviors on these platforms, suggesting that even extreme inertia mismatch can be overcome with sufficient data. However, it remains unclear whether high inertia imposes a hidden cost on learning, potentially worsening scaling with data or lowering the performance ceiling of modern robot learning methods. We investigate this question through controlled simulation experiments that vary robot and object masses. We find that reinforcement learning succeeds across a broad range of mass combinations, while behavior cloning learns more reliable policies when a lighter robot manipulates a heavier object. Heavier robots require more demonstrations to attain the same performance as lighter robots. Finally, because new hardware is rarely an option, we explore ways to lower an existing arm's inertia at contact with existing hardware, either by choosing a lower-inertia workspace before collecting data or by steering the demonstrator toward lower-inertia configurations during the task. In both cases, the lower-inertia demonstrations are imitated more reliably. Project website: \href{https://younghyopark.me/inertia-mismatch/}{https://younghyopark.me/inertia-mismatch/}
\end{abstract}

\section{Introduction}
\begin{figure}[t]
 \centering
 \includegraphics[width=\columnwidth,trim={0 8bp 0 0},clip]{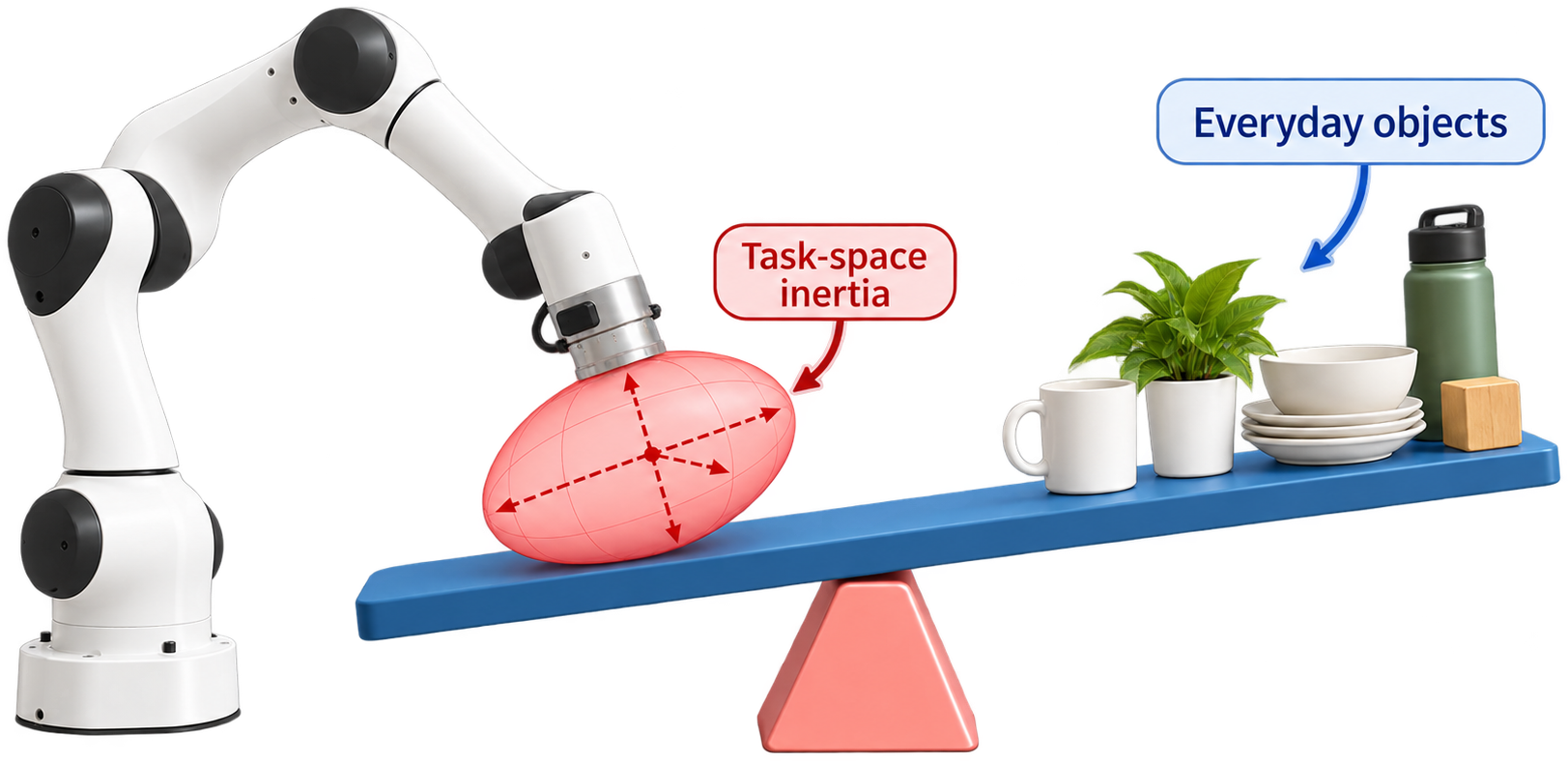}
 \par\nointerlineskip
 \setlength{\abovecaptionskip}{7pt}
 \caption{\textbf{Robot arms can significantly outweigh the everyday objects they manipulate.}
 A natural question arises: can modern robot learning fully compensate for this mismatch given sufficient data, or does it impose a
 hidden cost on learning reliable manipulation?}
 \label{fig:teaser}
\end{figure}

Conventional robot arms are extremely heavy~\cite{zinn2004actuation,kim2015lowinertia}
compared with many of the everyday objects they manipulate (Fig.~\ref{fig:teaser}).
This disparity is evident in public manipulation
demonstrations~\cite{mandlekar2022robomimic,mandlekar2023mimicgen}: we find that
arms present a mean effective mass of 6 to 16\,kg at contact, while the objects
weigh only 8 to 250\,g (Sec.~\ref{sec:background}).
As a result, even a small accidental bump from a heavy arm can knock a light object
out of place, whereas human arms can present substantially lower effective mass at
the hand~\cite{shin2010humanfriendly}.

Despite this substantial robot--object inertia mismatch, modern learning
methods have demonstrated a wide range of successful
manipulation behaviors on these platforms, from contact-rich pushing to precise
assembly~\cite{luo2024hilserl,chi2023diffusionpolicy}. These achievements can make
the robot's dynamics---especially this inertia mismatch---appear to be something
that enough data and learned task-level feedback can compensate for.

But this perspective may conceal the cost of compensating for those dynamics.
High inertia, especially at critical contact points, could make manipulation less forgiving of imperfect
actions, placing greater demands on the policy's precision and feedback.
The robot's dynamics may therefore influence how much data is needed to
achieve reliable behavior, and how much further performance improves as
learning is scaled. Although low-inertia hardware has been extensively
studied for safety and control performance~\cite{zinn2004actuation,kim2015lowinertia},
a central question remains for modern robot learning: \textbf{could high robot inertia be a hidden bottleneck to learning reliable manipulation skills?}

In this paper, we investigate this question through the lens of two prominent
robot learning paradigms: reinforcement learning (RL) and behavior cloning (BC).
Controlled simulation experiments reveal three findings:
\begin{enumerate}
 \item \textbf{RL learns successful manipulation across a broad range of inertia conditions.}
A trained policy achieves high success for robots both lighter and much
heavier than the objects they manipulate, demonstrating that RL can
accommodate substantial inertia mismatch. (Sec.~\ref{sec:rl})
\item \textbf{High robot inertia carries a measurable cost for BC.}
With the controller's free-space response matched across masses, BC becomes less reliable as the robot grows heavier relative to the object. Heavier robots also require more demonstrations to attain the same closed-loop performance as lighter robots. (Sec.~\ref{sec:bc})
\item \textbf{Demonstrations collected at lower inertia lead to better closed-loop
policies.} Task-space inertia depends on configuration and not only on the robot's
mass, so the cost of inertia mismatch can be reduced without changing the
hardware, by placing the workspace where the arm presents less inertia or by
steering the demonstrator toward lower-inertia configurations.
(Secs.~\ref{sec:design} and~\ref{sec:objective})
\end{enumerate}

Together, these findings identify an overlooked opportunity to improve robot
learning: robot--object inertia mismatch carries a measurable cost for BC even where RL still finds solutions, yet accounting for task-space inertia when setting up a robot or collecting its data can reduce that
cost and improve the reliability of learned manipulation skills.

\section{Background and Related Work}
\label{sec:background}
\myparagraph{Mass and Inertia of Articulated Robots.}
Mass measures resistance to linear acceleration through $F=ma$; rotational
inertia plays the corresponding role for rotation. For an articulated arm,
the mass matrix $M(q)$ relates joint accelerations to inertial torques at
configuration $q$. The Jacobian $J_v(q,c)$ maps joint motion to translational
motion at contact point $c$. Together they define the task-space inertia
matrix~\cite{khatib1995inertial}
\begin{equation}
 \Lambda_v(q,c)=\left[J_v(q,c)M(q)^{-1}J_v(q,c)^\top\right]^{-1},
 \label{eq:task_space_mass}
\end{equation}
wherever the inverse exists. This matrix describes the arm's resistance to
acceleration at that point; along a given direction, it reduces to a single
effective mass. The same arm can therefore present different
effective masses across postures and contact directions, even though its
total physical mass stays fixed.

\vspace{-6pt}
\begin{insetparagraph}
\begin{wrapfigure}{r}{0.50\columnwidth}
 \vspace{-\intextsep}
 \centering
 \includegraphics[width=\linewidth]{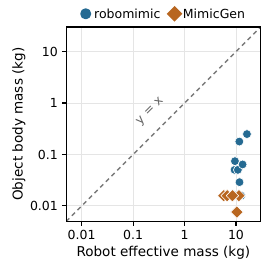}
 \par\nointerlineskip
 \setlength{\abovecaptionskip}{2pt}
 \caption{\textbf{Robots far outweigh objects in public demonstrations.} Mean masses; dashed line: \mbox{$y=x$}.}
 \label{fig:public_demonstration_mass_audit}
 \vspace{4pt}
\end{wrapfigure}
\myparagraph{Robot--Object Inertia Mismatch in Robot Datasets.}
We audit human and programmatically generated demonstrations from public simulation
datasets~\cite{mandlekar2022robomimic,mandlekar2023mimicgen} to assess the extent of robot--object inertia mismatch.
Across the audited tasks and arm models, mean robot effective mass at contact
substantially exceeds object mass (Fig.~\ref{fig:public_demonstration_mass_audit}).
The prevalence of this mismatch in these demonstrations motivates understanding
how it affects policy learning and the reliability of learned manipulation
behaviors.
\end{insetparagraph}

\myparagraph{Low-Inertia Robot Hardware.}
In traditional manipulation based on preprogrammed motion sequences,
limited task-level adaptation makes unexpected contact especially
challenging. When an obstacle interrupts a prescribed motion, high effective
inertia can generate large impact loads before feedback responds~\cite{zinn2004actuation}. Low moving
inertia and mechanical compliance provide physical tolerance to such
disturbances without requiring corrective commands, motivating robot designs
that make contact safer and more forgiving~\cite{zinn2004actuation,shin2010humanfriendly}.
Examples include tendon or cable transmissions that keep heavy actuators
near the base to reduce moving inertia~\cite{kim2015lowinertia}, and compliant
pneumatic muscles combined with lightweight links~\cite{shin2010humanfriendly}.

In the meantime, modern learned policies have shown that they can adjust their actions from sensory feedback and recover from
disturbances, enabling successful contact-rich manipulation on conventional
arms~\cite{luo2024hilserl,chi2023diffusionpolicy}. Such successes hint that
specialized low-inertia hardware might not be a prerequisite for performing these
tasks. However, they do not establish how difficult this feedback is to learn. Whether robot--object inertia mismatch increases the data needed
to learn reliable behavior or limits the performance attainable from that
data remains poorly understood.
\myparagraph{Controllers as Learning Interfaces.}
A controller is the interface through which a policy acts: position targets, learned
impedance, and gains all shape the resulting motion~\cite{aljalbout2024actionspace,martinmartin2019vices,bronars2026tune}.
Its tracking response controls free-space error sensitivity~\cite{bronars2026tune,seo2026behavior},
while policy frequency sets the time available for contact errors to develop before replanning.
Holding this interface fixed across robot masses lets us study the effect of physical inertia behind it.

\section{A Controlled Study on a Point-Mass Robot}
\label{sec:experiments}
\label{sec:planar}
\label{sec:planar_results}
\begin{figure}[!htb]
 \centering
 \vspace{-10pt}
 \includegraphics[width=\columnwidth]{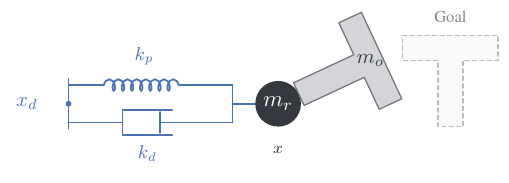}
 \vspace{-25pt}
 \caption{\textbf{A mass-scaled spring and damper drive a point-mass robot that pushes a T.}
 A circular robot with mass $m_r$ pushes an object with mass $m_o$ toward the outlined goal.
 The policy commands a position $x_d$, which the robot tracks through a virtual spring $k_p$ and damper $k_d$.
 These gains scale with $m_r$, so that every robot has the same unsaturated free-space
 response for a given natural frequency $\omega_n$ and damping ratio $\zeta$.}
 \label{fig:planar_schematic}
\end{figure}

To vary a robot's inertia directly, we first study robot--object inertia mismatch
in a controlled planar T-pushing environment
(Fig.~\ref{fig:planar_schematic}). A circular point-mass robot
of mass $m_r$ pushes a T-shaped object of mass $m_o$ toward a target pose.
In this baseline task, the T slides without floor friction against linear and
angular drag proportional to its mass.
We vary both masses from $0.1$ to $5$\,kg, covering different relative and
absolute mass scales. Arrival success requires the T to
reach 90\% target-footprint overlap before the episode ends.

\myparagraph{Controller Setup.}
The controller defines the policy's input--output interface: how commanded
position targets translate into robot motion. To isolate inertia's effect
during contact, this interface must have the same nominal free-space
response across robot masses. Matching the numerical gains alone would
not achieve this, because the response depends on the gains relative to
mass.
The policy commands a robot position $x_d$, held between updates, which a
faster PD controller tracks:
\begin{equation}
 m_r\ddot{x}=F+F_{\mathrm{contact}},\qquad
 F=k_p(x_d-x)-k_d\dot{x},
 \label{eq:controller}
\end{equation}
where $x$ is the actual position and $k_p,k_d$ are stiffness and damping
gains. At each tested natural frequency $\omega_n$ and damping ratio
$\zeta$, we set
\begin{equation}
 k_p=m_r\omega_n^2,\qquad k_d=2\zeta m_r\omega_n.
 \label{eq:gains}
\end{equation}

\begin{figure}[!t]
 \centering
 \includegraphics[width=\columnwidth]{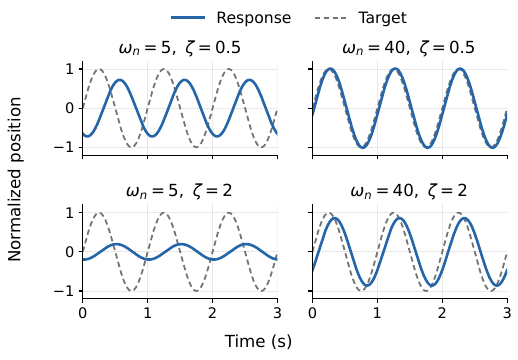}
 \vspace{-22pt}
 \caption{\textbf{Mass-scaled gains give each controller setting one free-space response at every robot mass.}
 Each panel shows the analytic steady-state response to a 1\,Hz sinusoidal target:
 columns vary natural frequency $\omega_n$ ($5$ and $40$\,rad/s), and rows vary
 damping ratio $\zeta$ ($0.5$ and $2$).}
 \label{fig:controller_response_corners}
\end{figure}

Heavier robots thus receive proportionally higher stiffness and damping.
This gives identical unsaturated free-space motion for the same target
history and initial state, while preserving differences in contact dynamics.
Because the controller itself shapes imitation~\cite{bronars2026tune}, we repeat
the following study over natural frequencies $\omega_n \in [5,40]$\,rad/s and damping
ratios $\zeta \in [0.5,2]$, using three logarithmically spaced values for each
(Fig.~\ref{fig:controller_response_corners}).

\subsection{Inertia Mismatch and Reinforcement Learning}
\label{sec:rl}

To study how robot--object inertia mismatch affects RL's ability to discover
successful manipulation strategies, we train a single dynamics-conditioned expert
with PPO~\cite{schulman2017ppo} jointly across randomized robot and object masses
and controller settings, which it receives as inputs. So that every robot can command
the same accelerations, we bound the expert's actions by a fixed acceleration scale
$a_{\max}$, and thus its commanded force by $F_{\max}=m_r a_{\max}$. We also vary the commanded
finish time $t_f$ from 2 to 4\,s.

\begin{myresult}[res:rl_competence]{RL Solution Existence}
RL discovers competent manipulation strategies across the tested
robot--object mass settings, including when the robot is much heavier
than the object.
\end{myresult}
\begin{insetparagraph}
\vspace{-7pt}
\begin{wrapfigure}{r}{0.46\columnwidth}
 \vspace{-\intextsep}
 \centering
 \includegraphics[width=\linewidth]{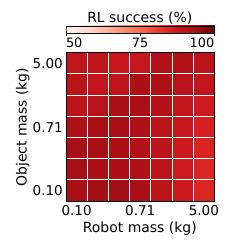}
 \par\nointerlineskip
 \setlength{\abovecaptionskip}{2pt}
 \caption{RL succeeds across the whole mass grid.}
 \label{fig:rl_mass_inset}
\end{wrapfigure}
Fig.~\ref{fig:rl_mass_inset} shows this competence across the mass grid, and it persists at both tested finish times, 2 and 4\,s, and across controller settings. Unlike BC, RL trains on data generated by its own exploration, which allows it to learn behaviors that compensate for the robot's inertia if the task is physicall possible to achieve.
\end{insetparagraph}

\subsection{Inertia Mismatch and Behavior Cloning}
\label{sec:bc}

Given that the RL succeeds across the mass grid, we now ask how reliably BC can imitate them. For each combination, we train a separate BC policy trained with 200 demonstrations, generated by the RL expert.

\begin{figure*}[t]
 \centering
 \begin{minipage}{1.00\textwidth}
 \begin{minipage}[t]{0.349\linewidth}
 \vspace{0pt}\centering
 \subfigure[T-pushing\label{fig:planar_bc}]{%
   \includegraphics[width=\linewidth]{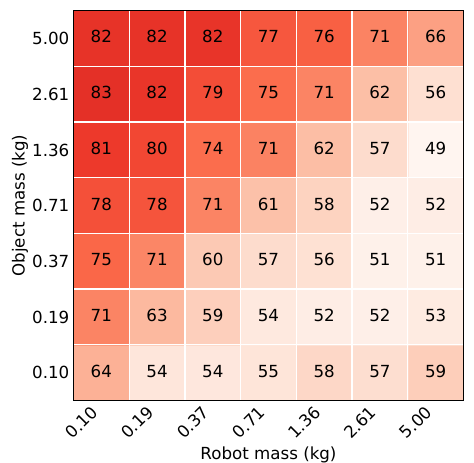}}
 \end{minipage}\hfill%
 \begin{minipage}[t]{0.578\linewidth}
 \vspace{0pt}\centering
 \subfigure[Table friction\label{fig:planar_friction}]{%
   \includegraphics[width=0.322\linewidth]{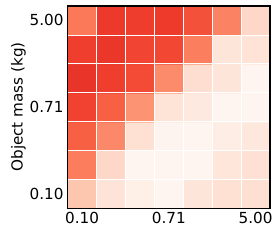}}\hfill%
 \subfigure[L-shaped object\label{fig:planar_l_object}]{%
   \includegraphics[width=0.322\linewidth]{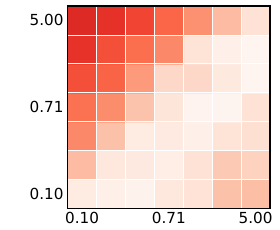}}\hfill%
 \subfigure[Large pusher: T\label{fig:planar_large_t}]{%
   \includegraphics[width=0.322\linewidth]{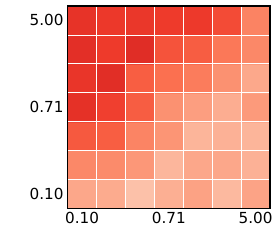}}
 \par\vspace{2pt}
 \subfigure[Large pusher: L\label{fig:planar_large_l}]{%
   \includegraphics[width=0.322\linewidth]{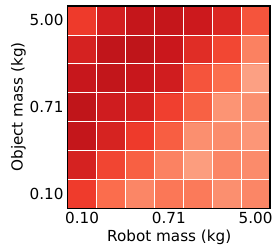}}\hfill%
 \subfigure[T-pushing, $t_f=2$ s\label{fig:planar_t2}]{%
   \includegraphics[width=0.322\linewidth]{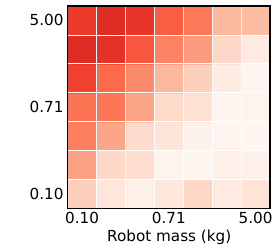}}\hfill%
 \subfigure[T-pushing, $t_f=4$ s\label{fig:planar_t4}]{%
   \includegraphics[width=0.322\linewidth]{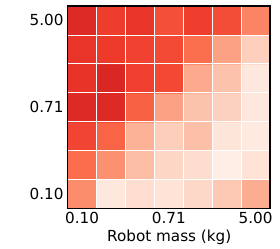}}
 \end{minipage}\hfill%
 \begin{minipage}[t]{0.065\linewidth}
 \vspace{0pt}\centering
 \includegraphics[width=\linewidth]{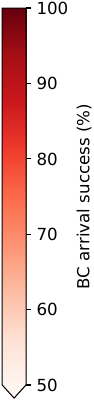}
 \end{minipage}
 \end{minipage}
 \caption{\textbf{BC, unlike RL, strongly prefers a lighter robot manipulating a heavier object.}
 Each
 policy is a Transformer trained on 200 demonstrations:
 from four steps of robot and object observations, it predicts five absolute planar
 position targets and executes one before replanning ($H=4$, $P=5$, $E=1$). All panels share mass axes and a color scale that saturates below 50\%. Cells show
 arrival success at each run's best closed-loop evaluation. Panels (a--e) average the
 settings common to all variants; (f--g)
 split the baseline by finish time.}
 \label{fig:planar}
\end{figure*}

\begin{myresult}[res:relative_mass]{Effect on BC}
With matched free-space response, BC performs best when a lighter robot manipulates a heavier object (i.e., the upper-left region of Fig.~\ref{fig:planar_bc}) and degrades severely as the robot grows heavier relative to the object.
\end{myresult}
Fig.~\ref{fig:planar_bc} shows this trend, averaged over controller settings and finish times. The cost follows the mass ratio rather than either mass alone: cells on a diagonal of the map, which share $m_r/m_o$, read about
the same. The preference also persists
when table friction replaces drag (Fig.~\ref{fig:planar_friction}), with
other contact geometries (Figs.~\ref{fig:planar_l_object}--\ref{fig:planar_large_l}),
and at either finish time (Figs.~\ref{fig:planar_t2}--\ref{fig:planar_t4}).

Before explaining this preference, we check three simpler explanations: too little data, softer gains, and poorer fitting.

\begin{figure}[t]
 \centering
 \includegraphics[width=\columnwidth]{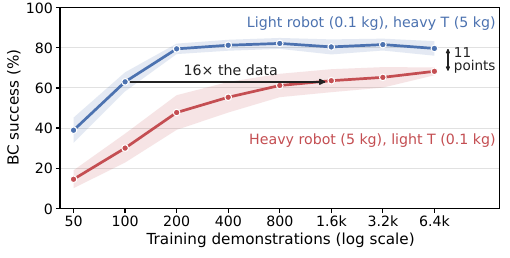}
 \par\nointerlineskip
 \setlength{\abovecaptionskip}{2pt}
 \caption{\textbf{More data narrows the gap but does not close it.} Curves show the mass-ratio extremes of Fig.~\ref{fig:planar_bc} over 18 settings with 95\%
 intervals; policies train for 50k updates, and success averages the last five
 closed-loop evaluations.}
 \label{fig:planar_data}
\end{figure}

\myparagraph{Beyond More Data.}
Because every policy in Fig.~\ref{fig:planar_bc} trains on only 200 demonstrations, the
gap between the map's light-robot and heavy-robot corners could simply reflect too
little data. However, more data narrows the gap without closing it
(Fig.~\ref{fig:planar_data}): a heavy robot pushing a light T needs 16 times the
demonstrations to match a light robot pushing a heavy T with 100, and even the
largest dataset leaves it behind.

\myparagraph{Beyond Free-Space Gains.}
Lighter robots also receive smaller physical gains in Eq.~\eqref{eq:gains}, and
compliant gains have been reported to benefit BC~\cite{bronars2026tune}. One might
therefore expect softer gains to explain their advantage. For zero-mean white-noise
position-target errors, however, free-space analyses give the stationary
position-error variance~\cite{bronars2026tune,seo2026behavior}
\begin{equation}
 \operatorname{Var}[\delta x]
 =\sigma^2\frac{k_p}{2k_d}
 =\sigma^2\frac{\omega_n}{4\zeta},
 \label{eq:gain_sensitivity}
\end{equation}
where $\sigma^2$ denotes the target-noise intensity. Substituting our
matched gains cancels $m_r$ between $k_p$ and $k_d$, leaving the sensitivity
factor independent of robot mass. Thus, smaller absolute gains do not give lighter robots lower free-space
error sensitivity. The map agrees: within a column of Fig.~\ref{fig:planar_bc}, the
robot and its physical gains are fixed, yet success generally rises with object
mass.

\vspace{-6pt}
\begin{insetparagraph}
\begin{wrapfigure}{r}{0.48\columnwidth}
 \vspace{-\intextsep}
 \centering
 \includegraphics[width=\linewidth]{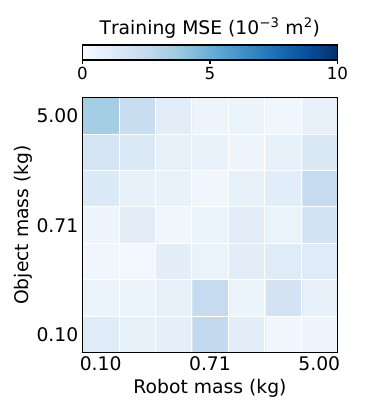}
 \par\nointerlineskip
 \setlength{\abovecaptionskip}{2pt}
 \caption{\textbf{Training error does not follow the mass preference.} Mean training MSE
 before each run's best closed-loop evaluation, averaged over the settings in Fig.~\ref{fig:planar_bc}.}
 \label{fig:planar_mse}
\end{wrapfigure}
\myparagraph{Beyond Fitting Error.}
The advantage could also reflect a poorer fit to the heavy robot's demonstrations.
However, the training-MSE map (Fig.~\ref{fig:planar_mse}) does not reproduce the
strong mass-ratio trend in closed-loop success, so the lighter-robot advantage
cannot be explained by more accurate fitting of demonstration actions. This suggests
that robot--object dynamics influence how policy errors translate into task
outcomes.
\par\vspace{20pt}
\end{insetparagraph}

\myparagraph{Inertia Mismatch and Error Propagation.}
Since neither the amount of data, the free-space gains, nor the fitting error
explains the advantage of lighter robots, we now examine why the trend in
Fig.~\ref{fig:planar} arises. Consider a
short interval in which the robot and object remain in rigid contact and
translate together along the push direction. Neglecting drag, compare two rollouts from the same state whose held targets
differ by a small error $e=\delta x_d$. Applying $F=ma$ to the combined mass gives
\begin{equation}
 (m_r+m_o)\delta\ddot{x}+k_d\delta\dot{x}+k_p\delta x=k_p e.
 \label{eq:contact_error}
\end{equation}
Here $\delta x$ is the shared robot and object displacement error.
Substituting the matched gains and writing $r=m_r/m_o$ yields the initial
object acceleration error
\begin{equation}
 \delta\ddot{x}(0)=\alpha\omega_n^2 e,\qquad
 \alpha=\frac{m_r}{m_r+m_o}=\frac{r}{1+r}.
 \label{eq:contact_ratio}
\end{equation}
The same target error therefore produces a larger initial object
acceleration as the robot-to-object mass ratio increases, even though
free-space tracking is identical. Because $\alpha$ depends only on $r$, this
sensitivity matches the pattern in Fig.~\ref{fig:planar_bc}: success changes little
along each diagonal, falls as $\alpha$ grows, and levels off once the robot is
several times heavier and $\alpha$ saturates.

\begin{figure*}[t]
 \centering
 \subfigure[Articulated arm\label{fig:rotor_arm}]{%
   \includegraphics[width=0.49\textwidth]{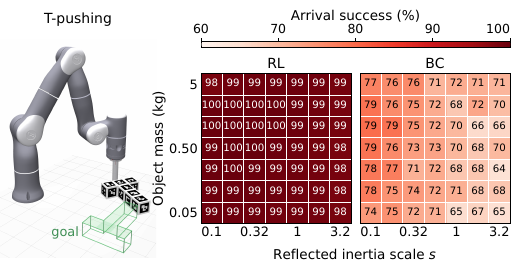}}\hfill%
 \subfigure[Dexterous hand\label{fig:rotor_hand}]{%
   \includegraphics[width=0.49\textwidth]{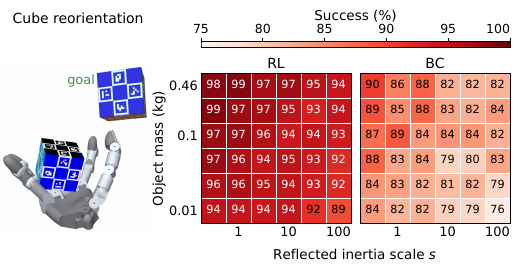}}
 \par\nointerlineskip
 \setlength{\abovecaptionskip}{6pt}
 \caption{\textbf{BC success falls as a robot reflects more motor inertia, at the same
 weight.} For each robot: its task, the RL expert's success,
 and BC's success with 200 demonstrations (three data draws per cell), over the
 reflected-inertia scale $s$ and the object's mass; $s=1$ is the unscaled robot. }
 \label{fig:rotor_inertia}
 \vspace{10pt}
\end{figure*}

\myparagraph{Policy Frequency and Error Propagation.}
A lower-frequency policy holds each position target for longer, which gives this acceleration error time to act. Let
$\Delta$ be the interval for which the policy holds its target. For a
short hold, Eq.~\eqref{eq:contact_error} gives
\begin{equation}
 \delta x(\Delta)=\tfrac12\alpha\omega_n^2 e\Delta^2
 +O(\Delta^3).
 \label{eq:contact_hold}
\end{equation}
Relative inertia sets the initial sensitivity, while the target hold
determines how long the error acts before the policy can revise it.
This effect is transient: a permanently held target still settles at the same
position for any mass. In manipulation, however, the
transient is what counts, because an unintended push can rotate the object, shift
contact, or separate the robot from it before the next update.
This transient offers a candidate mechanism for the preference for lighter robots.

\section{From a Point Mass to Articulated Robots}
\label{sec:hardware}
Sec.~\ref{sec:planar_results} studies a deliberately abstract robot: a point
mass, whose inertia is a single number $m_r$. An articulated arm's
inertia, however, is not one number but a matrix over its joints that changes
with its configuration,
\begin{equation}
 M(q)=M_{\mathrm{link}}(q)+\operatorname{diag}\big(N_i^2I_{\mathrm{rot},i}\big),
 \label{eq:mass_split}
\end{equation}
which adds the motors' reflected inertia to the links' own: a rotor of inertia
$I_{\mathrm{rot},i}$ spins $N_i$ times faster than its joint, so its inertia appears $N_i^2$ times larger
there. On a highly geared arm, this term is therefore a large share of $M(q)$;
on the Flexiv Rizon~4S model we use~\cite{menagerie2022github}, for instance, it is about half.

Given these differences, a natural question arises: would the cost of inertia
mismatch we identified using a point-mass robot carry over to an articulated arm as
well? To find out, we vary the arm's inertia as different gearing would, by
scaling only its reflected inertia, and repeat the mass study of
Sec.~\ref{sec:planar_results}. Here the arm's links, and with them
its weight, stay unchanged, so the same experiment also asks whether the cost
follows the arm's inertia rather than its weight.

\myparagraph{Controller Setup.}
Operational-space control (OSC)~\cite{khatib1987operational}
extends the matched-response comparison in Sec.~\ref{sec:experiments}
to an articulated arm. It accounts for the configuration-dependent
task-space inertia when converting position errors into actuator torques.
With the same gains across this comparison, it gives the end effector the
same nominal unit-mass spring--damper response in free space, while
preserving differences in effective mass at contact. Policies command
absolute XY tip targets; OSC maintains the tool's height and orientation.

\myparagraph{Scaling the Reflected Inertia.}
We multiply every joint's reflected inertia by seven values of $s$ from 0.1 to
3.16 and cross them with seven T masses. Torque limits scale with each joint's
inertia, mirroring the mass-scaled expert bound of Sec.~\ref{sec:planar}, so the
arm's free-space tip response stays nearly matched, while the effective mass it
presents at contact grows about fivefold.

\Needspace{8\baselineskip}
\begin{myresult}[res:rotor_arm]{Articulated Arm}
The cost of high robot inertia carries over to an articulated arm: BC degrades as
the arm's reflected inertia grows, at every object mass, even though the arm's
weight never changes.
\end{myresult}
While RL can find near-perfect solutions in every cell, BC success falls steadily as the arm reflects more inertia, at every T mass (Fig.~\ref{fig:rotor_arm}).

\myparagraph{A Dexterous Hand.}
We also ask whether the cost appears on a very different robot, a dexterous hand. A hand's geared fingers reflect motor inertia as well, and this reflected inertia supplies about three quarters of their joints' inertia. We therefore sweep $s$ up to 100 on a simulated 20-joint hand reorienting a cube in its
palm~\cite{wujimjlab}, where every joint keeps the same free-space response at every $s$. Fig.~\ref{fig:rotor_hand} shows the same trend: RL finds competent solutions across the grid, while BC succeeds more often on a low-inertia hand.

\begin{figure*}[t]
 \centering
 \begin{minipage}{1.00\textwidth}
 \centering
 \subfigure[Workspace placements\label{fig:flexiv_workspace}]{%
   \includegraphics[width=2.40in,height=2.40in,keepaspectratio]{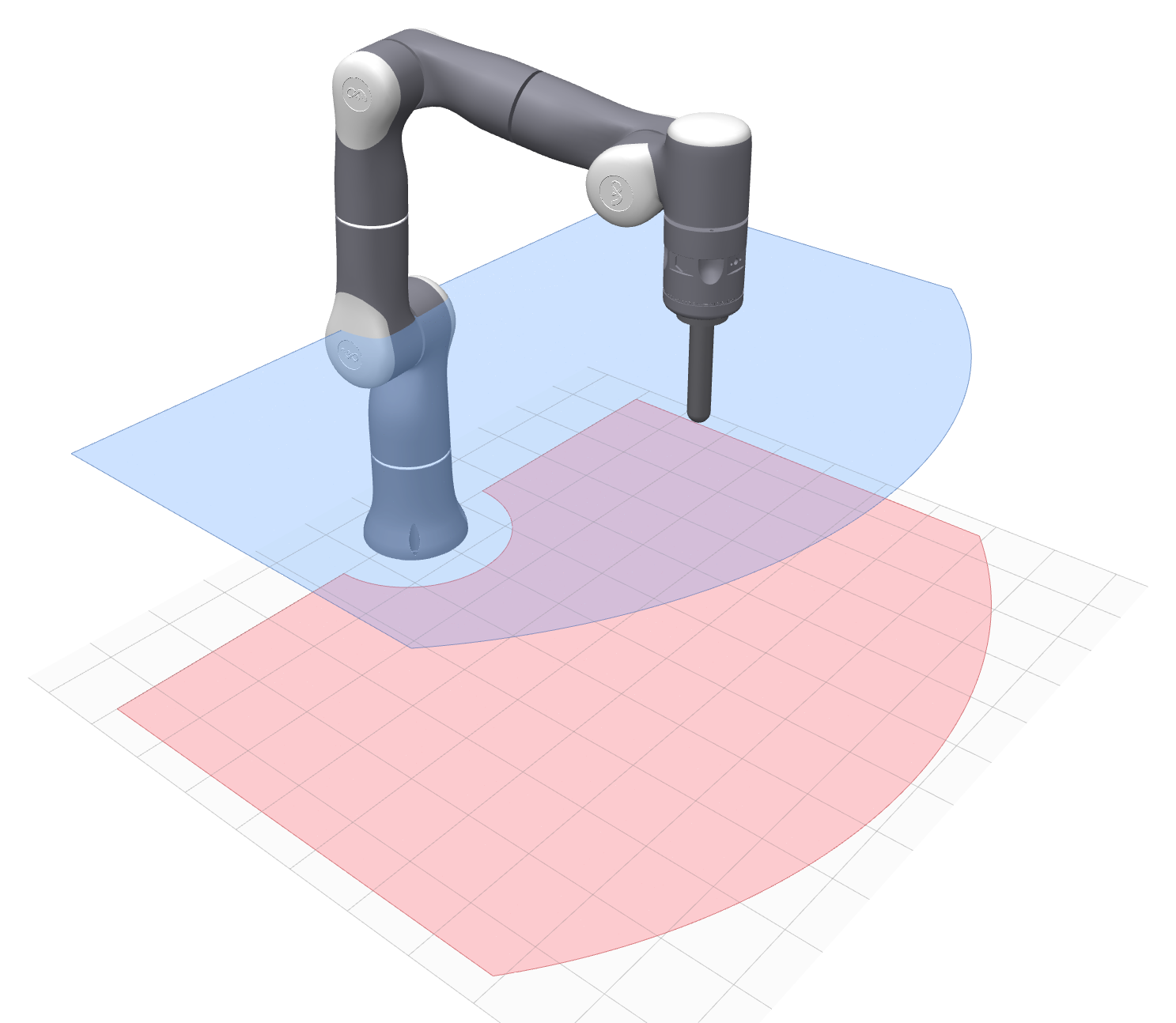}}\hfill%
 \subfigure[Baseline placement\label{fig:flexiv_ground_map}]{%
   \includegraphics[height=2.40in]{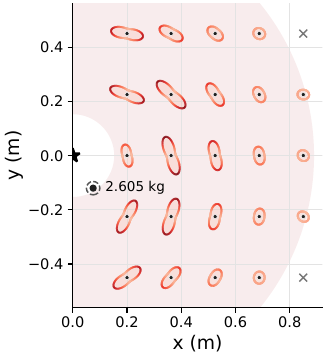}}\hfill%
 \subfigure[Lower-inertia placement\label{fig:flexiv_elevated_map}]{%
   \includegraphics[height=2.40in]{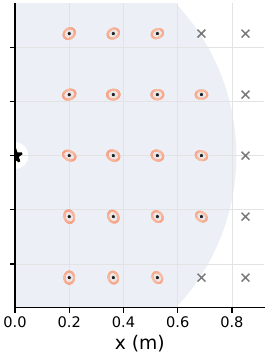}%
   \includegraphics[height=2.40in]{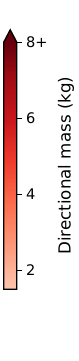}}
 \end{minipage}
 \caption{\textbf{Elevating the workspace lowers the inertia the same arm presents along planar push directions.}
 Each glyph plots the directional effective mass at one tip position over planar push
 directions: its radius and color both grow with mass, with identical scales in (b) and
 (c). The smaller glyphs in (c) identify a placement where the robot presents less inertia
 to planar contact, so such a map can reveal a lower-inertia workspace before collecting
 demonstrations or training a policy. The dashed circle
 is a radius key, gray crosses mark positions unreachable with the vertical tool, and
 the star marks the robot base. Red and blue sheets in (a) follow the mapped footprints
 at $z=0$ and $0.40$\,m; the gray grid marks the robot's base plane.}
 \label{fig:flexiv_glyphs}
 \vspace{18pt}
\end{figure*}

\section{Without Changing the Hardware}
\label{sec:practical_results}
Sec.~\ref{sec:hardware} shows that less reflected inertia improves imitation,
but changing hardware is rarely an option. Fortunately, the effective mass at contact is
not fixed by the hardware alone. At a contact point $c$ and along a contact
direction $u$, the arm presents a directional effective mass, the scalar
form of Eq.~\eqref{eq:task_space_mass},
\begin{equation}
 m(q,u)=\left[u^\top J_v(q,c)\,M(q)^{-1}J_v(q,c)^\top u\right]^{-1},
 \label{eq:directional_mass}
\end{equation}
the force needed per unit acceleration of the otherwise free arm along
$u$~\cite{khatib1995inertial}. Under ideal operational-space control, the free-space response is fixed, while a
contact force along $u$ initially acts against this mass~\cite{khatib1987operational}. Thus $m(q,u)$
takes the place of $m_r$ in the initial response of Eq.~\eqref{eq:contact_ratio}.
Because the Jacobian depends on posture, however,
the same arm can present very different masses to the same object
(Fig.~\ref{fig:flexiv_glyphs}).

The effective mass along a dataset's contacts therefore also depends on choices made around
a fixed robot. We make these choices explicit (Sec.~\ref{sec:formulation}) and
then study one instance of each resulting lever (Secs.~\ref{sec:design}
and~\ref{sec:objective}). Because each lever also changes the demonstrator's strategy,
these studies show that a lever works rather than isolate inertia as its cause.

\subsection{Setup and Demonstrator}
\label{sec:formulation}
\myparagraph{Effective Mass Along Demonstrations.}
On a fixed robot, two kinds of choices determine the effective masses along a
dataset's contacts. The \emph{setup} $\theta$ comprises everything other than
the arm itself that is fixed before data collection, such as the tool, where the task sits
relative to the base, and how the controller resolves redundancy.
With a task-space interface, $\theta$ maps each contact location $x$ to a
configuration $q_\theta(x)$ and thereby defines an inertia landscape over the
task region, \mbox{$m_\theta(x,u)=m(q_\theta(x),u)$}. The
\emph{demonstrator} $\pi$ decides which contacts occur, inducing a
distribution $p_\pi(x,u)$. Together, they set the mean effective mass along those contacts,
\begin{equation}
 \bar m(\theta,\pi)=\mathbb{E}_{(x,u)\sim p_\pi}\left[m_\theta(x,u)\right].
 \label{eq:encountered_inertia}
\end{equation}
With a joint-space interface, the demonstrator also chooses the posture, which
thus moves from $\theta$ to $\pi$.

\myparagraph{Two Levers.}
The mean $\bar m$ can therefore be lowered through either argument. \emph{Design-driven} interventions change the landscape before any
data is collected,
\begin{equation}
 \theta^\star=\arg\min_{\theta}\;\mathbb{E}_{\rho}\big[m_\theta(x,u)\big],
 \label{eq:design_lever}
\end{equation}
where $\rho$ covers the task region and plausible contact directions. They need
only the robot model. \emph{Objective-driven} interventions
instead keep $\theta$ and steer the demonstrator toward lower-inertia contacts
among the many ways to solve the task,
\begin{equation}
 \pi^\star=\arg\max_{\pi}\;J_{\mathrm{task}}(\pi)
 -\beta\,\mathbb{E}_{p_\pi}\big[h\big(m_\theta(x,u)\big)\big],
 \label{eq:objective_lever}
\end{equation}
with $J_{\mathrm{task}}$ the task objective, $h$ increasing, and $\beta>0$.
\vspace{3pt}

\subsection{Design-Driven: Choose the Setup Before Collecting Data}
\label{sec:design}
Eq.~\eqref{eq:design_lever} lets us compare candidate setups from inertia maps alone,
before any data exists. We vary a single design
variable, the height of the workspace, on T-pushing with the Flexiv arm.
As Fig.~\ref{fig:flexiv_glyphs} shows,
elevating the workspace lowers the inertia the arm presents along planar push
directions. We therefore compare a
baseline placement on the ground ($z=0$) with a lower-inertia placement elevated to
$z=0.40$\,m (Fig.~\ref{fig:flexiv_workspace}).

\begin{figure}[!htb]
 \centering
 \includegraphics[width=0.98\columnwidth]{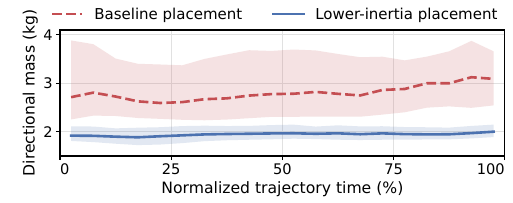}
 \vspace{-10pt}
 \caption{\textbf{Workspace placement lowers inertia along the demonstrated motions.}
 Curves show the median and interquartile range over 1,800 training trajectories per
 placement, aligned by normalized elapsed time. Mass is the directional effective mass
 of Eq.~\eqref{eq:directional_mass} along measured tip XY velocity, omitting
 near-stationary samples.}
 \label{fig:flexiv_expert_mass}
\end{figure}

\myparagraph{RL Experiment.}
We train
separate PPO experts at each placement with multiple random
seeds and collect successful rollouts as BC demonstrations. Following the same conclusion we had for previous sections, RL solves
the task at both placements with similar success. Along the demonstrated behaviors, the baseline placement has persistently higher
effective mass (Fig.~\ref{fig:flexiv_expert_mass}). The
inertia difference therefore carries into BC's training data.

\myparagraph{BC Experiment.}
For each placement, we train BC policies on these demonstrations, varying
observation-history length $H$, prediction length $P$, and execution length $E$ so that the comparison does not rest on one BC training and evaluationsetting.

\begin{myresult}[res:workspace_placement]{Design: Workspace Height}
Placing the task in a lower-inertia workspace improves imitation while
keeping the robot, action interface, and nominal controller gains fixed.
\end{myresult}
\begin{insetparagraph*}
\begin{wrapfigure}{r}{0.46\columnwidth}
 \centering
  \vspace{-3pt}
 \includegraphics[width=\linewidth]{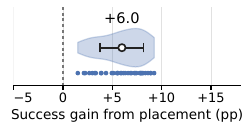}
 \par\nointerlineskip
 \setlength{\abovecaptionskip}{2pt}
 \caption{\textbf{All 36 settings gain from the lower-inertia placement.} Dots: settings, averaged over expert seeds and data draws; marker and bar: mean $\pm$ SD at 150k updates; pp: percentage points.}
 \vspace{-3pt}
 \label{fig:workspace_sweep_summary}
\end{wrapfigure}
\insetfigurefits
The gain holds in every evaluated setting
(Fig.~\ref{fig:workspace_sweep_summary}). It comes without a consistent change in action MSE
(Figs.~\ref{fig:flexiv_bc_curve}--\ref{fig:flexiv_mse}), so fitting error
does not explain it.
With the same arm at both placements, its weight and joint-space reflected inertia
are unchanged, so neither can explain this gain. Together with
Sec.~\ref{sec:hardware}, this points to the arm's effective mass at contact, which its
gearing and its posture both set, rather than to its weight.
\end{insetparagraph*}

\begin{figure*}[t]
 \centering
 \setlength{\practicalfigurecontentwidth}{\dimexpr\textwidth-0.08in\relax}
 \noindent\makebox[\textwidth][c]{%
 \begin{minipage}[b]{0.5\practicalfigurecontentwidth}
 \centering
 \subfigure[\strut T-pushing\label{fig:flexiv_task}]{%
   \begin{minipage}[t]{\linewidth}
    \begin{minipage}[t]{0.49\linewidth}
     \centering
     \includegraphics[trim=0 90bp 0 360bp,clip,width=\linewidth]{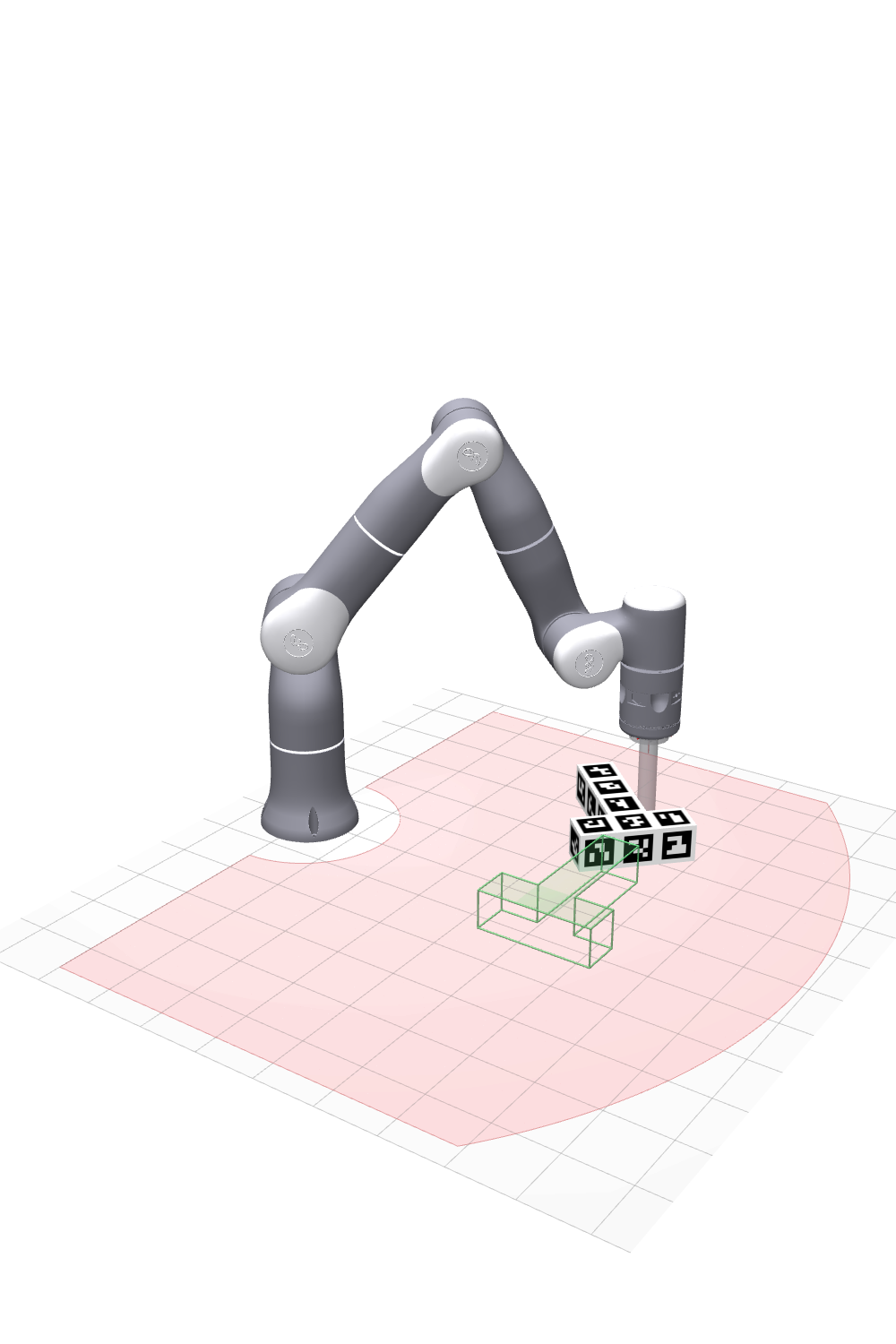}
     \par{\fontsize{7}{8}\selectfont Baseline\par}
    \end{minipage}\hfill%
    \begin{minipage}[t]{0.49\linewidth}
     \centering
     \includegraphics[trim=0 90bp 0 360bp,clip,width=\linewidth]{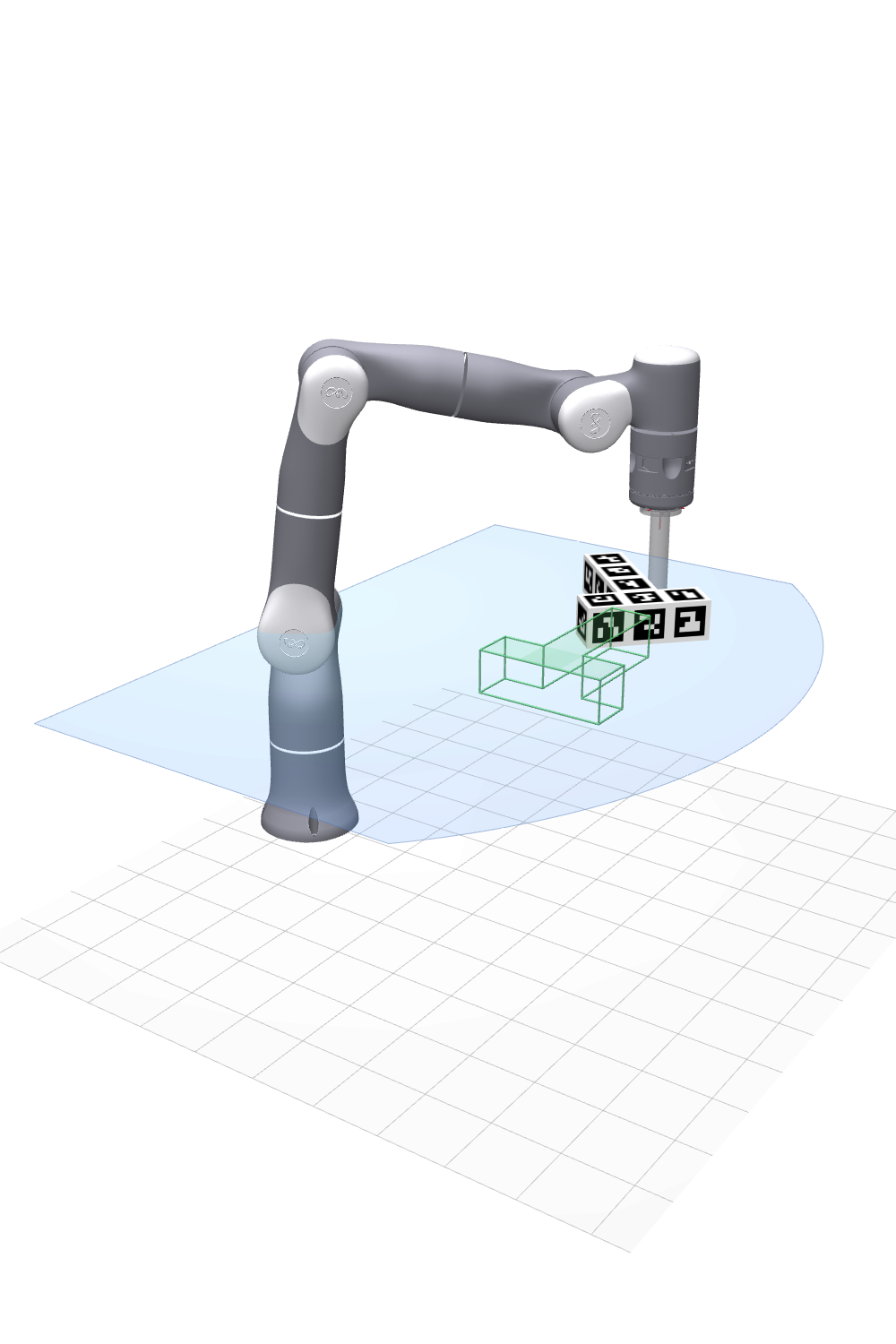}
     \par{\fontsize{7}{8}\selectfont Lower inertia\par}
    \end{minipage}
   \end{minipage}}
 \par\vspace{0pt}
 \end{minipage}\hspace{0.04in}%
 \begin{minipage}[b]{0.3\practicalfigurecontentwidth}
 \centering
 \subfigure[Rollout success\label{fig:flexiv_bc_curve}]{%
   \includegraphics[width=\linewidth]{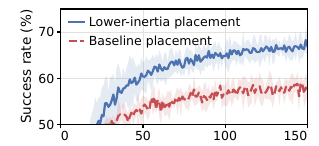}}
 \par\vspace{1pt}
 \subfigure[\strut Validation action MSE\label{fig:flexiv_mse}]{%
   \includegraphics[width=\linewidth]{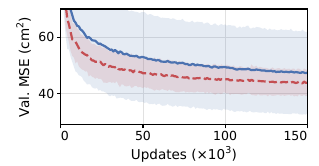}}
 \par\vspace{0pt}
 \end{minipage}\hspace{0.04in}%
 \begin{minipage}[b]{0.2\practicalfigurecontentwidth}
 \centering
 \subfigure[$H{=}2,\ P{=}1,\ E{=}1$\label{fig:flexiv_support1}]{%
   \includegraphics[width=\linewidth]{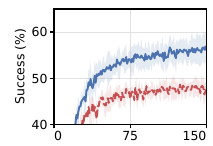}}
 \par\vspace{1pt}
 \subfigure[\strut $H{=}4,\ P{=}2,\ E{=}1$\label{fig:flexiv_support2}]{%
   \includegraphics[width=\linewidth]{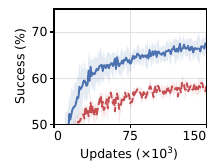}}
 \par\vspace{0pt}
 \end{minipage}}
 \par\nointerlineskip
 \setlength{\abovecaptionskip}{6pt}
 \caption{\textbf{Moving the task improves imitation without replacing the robot.}
 Higher rollout success in (b) comes without a consistent change in validation action
 error in (c), where the two placements overlap across experts. MLP policies train on 200 demonstrations to predict absolute XY tip
 targets from histories of tip position and object pose. Panels (b--e) show settings
 picked for large gains, with the largest, $H=8,P=4,E=1$, in (b--c);
 Fig.~\ref{fig:workspace_sweep_summary} shows all 36. Panel (a) shows the baseline
 (left, red) and lower-inertia (right, blue) placements from the same view;
 sheets follow the regions in Fig.~\ref{fig:flexiv_glyphs}, gray grids mark
 $z=0$, and green marks the goals. Curves average three
 experts and three data draws per expert; shading is sample SD across
 expert means.}
 \label{fig:flexiv_design}
\end{figure*}

\vspace{0pt plus 2pt}
\begin{myresult}[res:policy_frequency]{Policy Frequency}
The benefit of lower-inertia workspace placement moderately grows as policy
commands become less frequent.
\end{myresult}
\begin{insetparagraph}
\begin{wrapfigure}{r}{0.50\columnwidth}
 \vspace{-\intextsep}
 \centering
 \includegraphics[width=\linewidth]{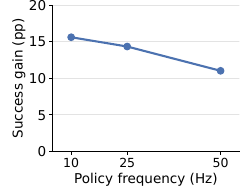}
 \par\nointerlineskip
 \setlength{\abovecaptionskip}{2pt}
 \caption{BC performance gap between two workspace settings across policy frequencies.}
 \label{fig:flexiv_policy_frequency}
\end{wrapfigure}
A longer policy interval gives a contact error more time to act
(Eq.~\eqref{eq:contact_hold}), so we next ask how this gain depends on policy
frequency. BC success falls at both placements at lower rates, while the
lower-inertia placement keeps its advantage (Fig.~\ref{fig:flexiv_policy_frequency}).
\end{insetparagraph}

\subsection{Objective-Driven: Steer the Demonstrator}
\label{sec:objective}
This lever keeps the setup and changes which contacts occur. We instantiate Eq.~\eqref{eq:objective_lever} with an
RL expert whose reward penalizes effective mass, on 6D cube
reorientation, in which a tool on a Franka arm pushes and tips a cube to a goal
pose. This expert acts in joint space, so the posture
becomes part of its choice $\pi$.

\myparagraph{Controller Setup.}
Here we relax the matched task-space response requirement and use
joint-space position control, with the same PD gains during demonstration
collection and BC execution. Fixed joint gains
allow the task-space response to vary with configuration, so this
experiment tests whether lower-inertia demonstrations also benefit
learning through a standard joint-position interface.

\begin{figure}[!tb]
 \centering
 \includegraphics[width=0.98\columnwidth]{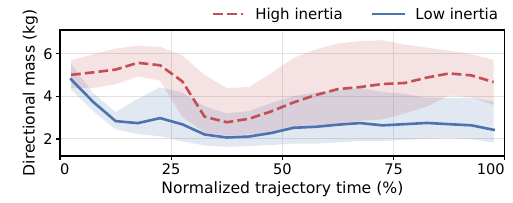}
 \vspace{-10pt}
 \caption{\textbf{Two RL experts solve the same task with distinct effective mass.}
 Median and interquartile range across 2,000 paired successful demonstrations per
 condition, from one expert each, aligned by normalized elapsed time. The reward and this plot
 evaluate $m$ along the tool--cube contact normal, during approach and contact.}
 \label{fig:franka_experts}
\end{figure}

\myparagraph{RL Experiment.}
We train two PPO experts with the same task reward and an additional
term favoring low or high effective mass $m$:
\begin{equation}
 r_{\mathrm{inertia}}=-\Delta t
 \begin{cases}
  \beta_{\mathrm{low}}\,\dfrac{m}{m+m_{\mathrm{ref}}}, & \text{low inertia},\\[10pt]
  \beta_{\mathrm{high}}\,\dfrac{m_{\mathrm{ref}}}{m+m_{\mathrm{ref}}}, & \text{high inertia}.
 \end{cases}
 \label{eq:inertia_reward}
\end{equation}
Here $m_{\mathrm{ref}}$ is a reference mass, $\Delta t$ is the policy interval, and
$\beta_{\mathrm{low}}$ and $\beta_{\mathrm{high}}$ set the penalty rates.
The low-inertia expert instantiates
Eq.~\eqref{eq:objective_lever} with $h(m)=m/(m+m_{\mathrm{ref}})$; the high-inertia expert
reverses the preference to provide a contrast. Both experts solve the task from nearly
all shared initial conditions, and the demonstrations we collect where both succeed
exhibit clearly different effective mass (Fig.~\ref{fig:franka_experts}).

\addtocounter{figure}{1}
\begin{figure*}[!t]
 \centering
 \setlength{\practicalfigurecontentwidth}{\dimexpr1.000\textwidth-0.08in\relax}
 \noindent\makebox[\textwidth][c]{%
 \begin{minipage}[t]{0.4\practicalfigurecontentwidth}
 \vspace{0pt}\centering
 \subfigure[6D cube reorientation\label{fig:franka_task}]{%
   \includegraphics[width=\linewidth,height=2.200in,keepaspectratio]{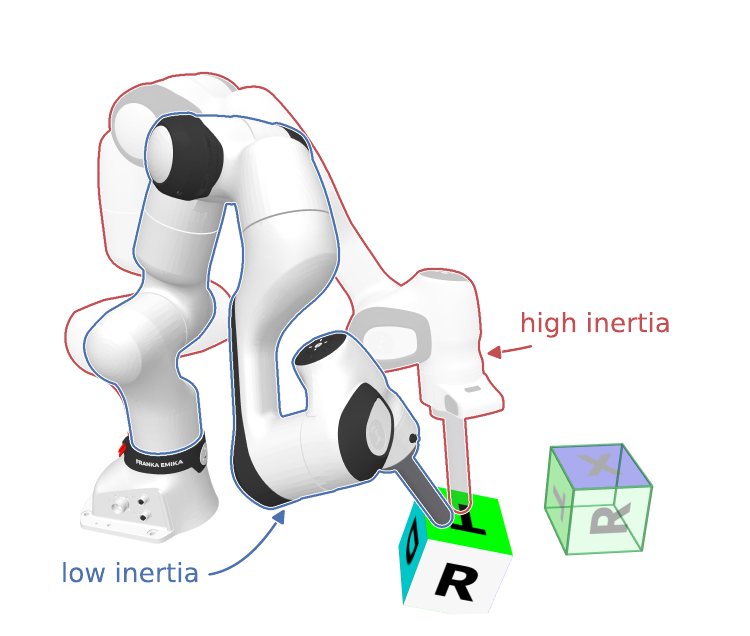}}
 \end{minipage}\hspace{0.04in}%
 \begin{minipage}[t]{0.4\practicalfigurecontentwidth}
 \vspace{0pt}\centering
 \subfigure[Rollout success\label{fig:franka_bc_curve}]{%
   \includegraphics[width=\linewidth]{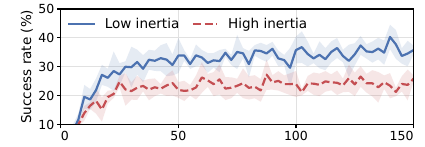}}
 \par\vspace{1pt}
 \subfigure[Training action MSE\label{fig:franka_mse}]{%
   \includegraphics[width=\linewidth]{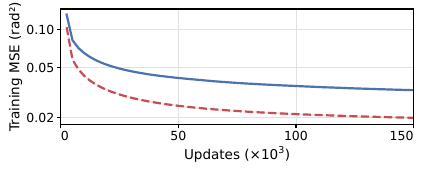}}
 \end{minipage}\hspace{0.04in}%
 \begin{minipage}[t]{0.2\practicalfigurecontentwidth}
 \vspace{0pt}\centering
 \subfigure[$H{=}20,\ P{=}1$\label{fig:franka_support1}]{%
   \includegraphics[width=\linewidth]{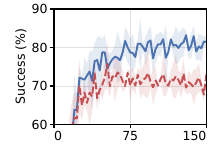}}
 \par\vspace{1pt}
 \subfigure[$H{=}20,\ P{=}10$\label{fig:franka_support2}]{%
   \includegraphics[width=\linewidth]{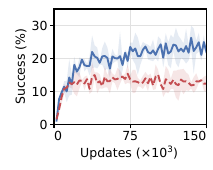}}
 \end{minipage}}
 \caption{\textbf{Joint-position BC trained on lower-inertia demonstrations succeeds more often.}
 BC trains on 2,000 demonstrations from the low- or high-inertia expert, with
 the same robot and position controller. Higher success in (b) accompanies larger action
 MSE in (c). Panels (b--e) show the settings repeated across seeds that gain most, with the largest, $H=5,P=10$, in (b--c);
 Fig.~\ref{fig:reward_sweep_summary} shows all 16 at 50k. Panel (a)
 overlays recorded expert frames at similar cube poses, showing distinct arm
 configurations; the green goal pose is
 illustrative. Shading is sample SD across
 learner seeds.}
 \label{fig:franka_reward}
\end{figure*}
\addtocounter{figure}{-2}

\myparagraph{BC Experiment.}
We train separate BC policies on the low- and high-inertia demonstrations.

\begin{myresult}[res:demonstration_reward]{Objective: Expert Reward}
Joint-position BC trained on lower-inertia demonstrations succeeds more often than on higher-inertia
ones, despite larger action MSE.
\end{myresult}
\begin{insetparagraph}
\vspace{-7pt}
\begin{wrapfigure}{r}{0.46\columnwidth}
 \vspace{-\intextsep}
 \centering
 \includegraphics[width=\linewidth]{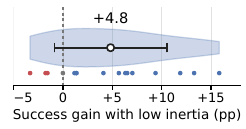}
 \par\nointerlineskip
 \setlength{\abovecaptionskip}{2pt}
 \caption{\textbf{Most settings gain from lower-inertia demonstrations.} Gain at 50k updates over 16 settings; marker and bar: mean $\pm$ SD.}
 \label{fig:reward_sweep_summary}
\end{wrapfigure}
Most settings of history and prediction length benefit
(Fig.~\ref{fig:reward_sweep_summary}), and training MSE is higher in all 16.
In the setting with the largest gain among those repeated across seeds,
lower-inertia demonstrations improve BC success in every seed despite
larger training MSE (Figs.~\ref{fig:franka_bc_curve}--\ref{fig:franka_mse}).
\end{insetparagraph}

\section{Conclusion and Remarks}
\label{sec:conclusion}
Robot--object inertia mismatch carries a measurable cost for learning
reliable manipulation through behavior cloning. In our controlled
simulations, RL finds successful strategies across a broad
range of masses, while BC favors a lighter robot relative to the object.
Heavier robots require more demonstrations to attain the same performance, and even the
largest dataset we tested leaves them behind. A similar cost appears on an articulated arm
and a dexterous hand when only their reflected inertia grows. Arms that reflect less
motor inertia, such as those with lower gear ratios, could therefore help. Because effective mass also depends on configuration, however, an existing arm can present less of it, either through a setup chosen before collecting data or through a demonstrator steered toward lower-inertia solutions. In one instance of
each, the lower-inertia condition gives more reliable closed-loop BC on the same
arm. These findings
also raise open questions.

\myparagraph{Retargeting Action-Less Demonstrations.}
Human videos and handheld demonstrations, including those collected with
UMI~\cite{chi2024umi}, do not directly specify the target robot's joint
configurations or the inertia it would present at contact. Human arms can
present much lower effective mass than conventional
robots~\cite{shin2010humanfriendly}, so reproducing a \mbox{demonstrated} tool
path need not reproduce its physical interaction with the object. This
suggests considering task-space inertia when retargeting demonstrations:
among configurations that realize the desired tool motion, retargeting
could favor lower inertia along relevant contact directions. Whether
this improves policy reliability and data efficiency is an open question.

\myparagraph{Teleoperation Interface Design.}
For data collected directly on a robot, an open question is how interface
design can encourage lower effective mass in demonstrations.
A teacher--follower system with a dynamically matched teacher arm could
let the operator feel inertial resistance and potentially favor
lower-inertia motions. Alternatively, inverse kinematics could
favor lower-inertia postures while tracking the operator's tool
commands.
Testing whether these interfaces improve learned policy performance
would extend these levers to human data collection.

\myparagraph{Effect on Sim-to-Real Transfer.}
Sim-to-real discrepancies in dynamics and sensing can push a policy away
from its training distribution~\cite{peng2018simtoreal,bronars2026tune}. The resulting
action errors may further alter contact and compound the state deviation,
suggesting a connection to error propagation in imitation
learning~\cite{pfrommer2022tasil}. Prior work on controller gains likewise
finds that small modeling errors can accompany large closed-loop transfer
errors~\cite{bronars2026tune}. Our findings motivate a complementary
hypothesis: lower robot inertia relative to the object may limit this
amplification during contact. Comparing low- and high-inertia configurations
under controlled dynamics and observation shifts, with matched nominal
controller response, could test whether the BC advantage extends to
sim-to-real transfer and complements dynamics randomization.

\section*{Acknowledgment}
This research was financially supported by the Ministry of Trade,
Industry, and Energy (MOTIE), Korea, under the ``Global Industrial
Technology Cooperation Center program'' supervised by the Korea Institute
for Advancement of Technology (KIAT) (Grant No. P0028435).
We used AI tools (Claude Code and OpenAI Codex) to assist with plotting code and
managing experimental runs on GPU clusters for
Secs.~\ref{sec:experiments} to~\ref{sec:practical_results}. The authors retain responsibility for the
scientific content and conclusions.

\newcommand{\BIBentryALTinterwordspacing}{\rightskip=0pt plus 4em\relax}
\newcommand{\BIBentrySTDinterwordspacing}{\par\rightskip=0pt\relax}
\bibliographystyle{IEEEtran}
\bibliography{references,opening_references}

\end{document}